\documentclass[10pt,twocolumn,letterpaper]{article}

\usepackage[pagenumbers]{cvpr} %

\definecolor{cvprblue}{rgb}{0.21,0.49,0.74}
\usepackage[pagebackref,breaklinks,colorlinks,allcolors=cvprblue]{hyperref}
\usepackage{multirow}
\usepackage{graphicx}
\usepackage{MnSymbol,wasysym}
\usepackage{algorithm}
\usepackage{algpseudocode}
\usepackage{amsmath}
\usepackage{array}
\usepackage{float}
\usepackage{placeins}
\usepackage[misc]{ifsym}
 \usepackage{graphicx}
\usepackage{fancyhdr}
\title{\vspace{-1cm}BlueLM-V-3B: Algorithm and System Co-Design for Multimodal Large Language Models on Mobile Devices}

\author{
Xudong Lu$^{*2,1\dagger}$, Yinghao Chen$^{*1}$, Cheng Chen$^{*1}$, Hui Tan$^{*1}$,
Boheng Chen$^{1}$,\\ Yina Xie$^{1}$, Rui Hu$^{1}$, Guanxin Tan$^{1}$, Renshou Wu$^{1}$, Yan Hu$^{1}$, Yi Zeng$^{1}$, Lei Wu$^{1}$, Liuyang Bian$^{1}$,\\ Zhaoxiong Wang$^{1}$, Long Liu$^{1}$, Yanzhou Yang$^{1}$, Han Xiao$^{2,1\dagger}$,\\  
Aojun Zhou$^{2}$, Yafei Wen$^{1}$, Xiaoxin Chen$^{1}$, Shuai Ren$^{1\,\ddagger\,{\textrm{\Letter}}}$, Hongsheng Li$^{2\,{\textrm{\Letter}}}$\\
\text{$^1$vivo AI Lab\quad $^2$CUHK MMLab}\\
\texttt{\{luxudong@link,hsli@ee\}.cuhk.edu.hk}\\
\texttt{shuai.ren@vivo.com}\thanks{$^*$Equal contribution $^{\textrm{\Letter}}$Corresponding author $^\ddagger$Project lead $^\dagger$Interns at vivo.\\}
}
\vspace{-6em}
\makeatletter
\def\thanks#1{\protected@xdef\@thanks{\@thanks
        \protect\footnotetext{\hspace{-2em}#1}}}
\makeatother
\begin{document}
\maketitle
\begin{abstract}
The emergence and growing popularity of multimodal large language models (MLLMs) have significant potential to enhance various aspects of daily life, from improving communication to facilitating learning and problem-solving. Mobile phones, as essential daily companions, represent the most effective and accessible deployment platform for MLLMs, enabling seamless integration into everyday tasks. However, deploying MLLMs on mobile phones presents challenges due to limitations in memory size and computational capability, making it difficult to achieve smooth and real-time processing without extensive optimization. In this paper, we present \textbf{BlueLM-V-3B}, an algorithm and system co-design approach specifically tailored for the efficient deployment of MLLMs on mobile platforms. To be specific, we redesign the dynamic resolution scheme adopted by mainstream MLLMs and implement system optimization for hardware-aware deployment to optimize model inference on mobile phones. BlueLM-V-3B boasts the following key highlights: (1) \textbf{Small Size}: BlueLM-V-3B features a language model with 2.7B parameters and a vision encoder with 400M parameters. (2) \textbf{Fast Speed}: BlueLM-V-3B achieves a generation speed of \textbf{24.4} token/s on the MediaTek Dimensity 9300 processor with 4-bit LLM weight quantization. (3) \textbf{Strong Performance}: BlueLM-V-3B  has attained the highest average score of \textbf{66.1} on the OpenCompass benchmark among models with $\leq$ 4B parameters and surpassed a series of models with much larger parameter sizes (e.g., MiniCPM-V-2.6, InternVL2-8B).
\end{abstract}

\vspace{-1em}
\section{Introduction}\label{intro}

Large language models (LLMs)~\cite{brown2020language,touvron2023llama1,touvron2023llama2} have gained significant attention due to their potential to solve various complex tasks~\cite{wei2022emergent, trinh2024solving}. Multimodal large language models (MLLMs) extend the capabilities by processing and integrating various forms of data—such as text, images, and audio—enabling richer interaction and a deeper understanding of context, which lead to more intuitive user experiences~\cite{achiam2023gpt,chen2024internvl,chen2024far,bai2023qwen,wang2024qwen2,mckinzie2024mm1,zhang2024mm15methodsanalysis,tong2024cambrian}. As research and applications of LLMs and MLLMs continue to evolve, more studies are exploring the feasibility of deploying the models on a variety of devices, including cloud-based platforms~\cite{wang2024cloud}, desktop PCs~\cite{song2023powerinfer}, and even edge devices~\cite{ding2024enhancing,qu2024mobile,hua2024interactivespeculativeplanningenhance,yao2024minicpm,xue2024powerinfer}. This trend emphasizes the need for optimizing model performance and resource efficiency to ensure accessibility across diverse platforms.

\begin{figure}
    \centering
    \vspace{-2em}
    \includegraphics[width=\linewidth]{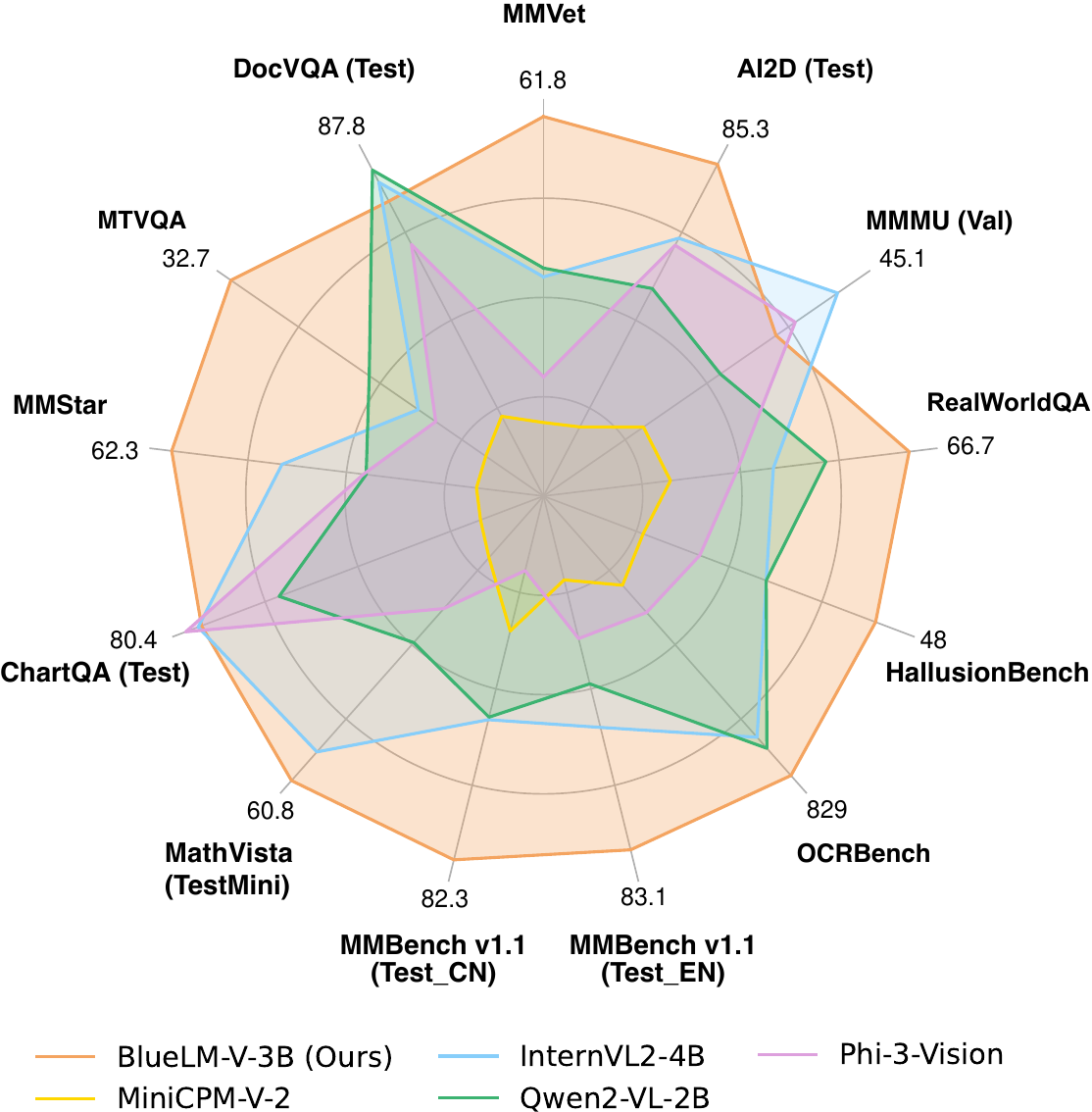}
    \vspace{-1.5em}
    \caption{\textbf{Comparison with mainstream MLLMs.} We compare the performance of several mainstream MLLMs with a parameter count similar to that of BlueLM-V-3B across multiple benchmarks. BlueLM-V-3B leads in the majority of datasets.}
    \label{fig:radar_fig}
    \vspace{-1em}
\end{figure}

Among available platforms, mobile phones stand out as the most efficient and accessible tool for deploying MLLMs. {Firstly}, it enables real-time, on-device processing, allowing users to interact with the model offline. This enhances privacy and reduces latency~\cite{qu2024mobile,ding2024enhancing}. {Secondly}, mobile deployment improves accessibility, allowing users to leverage advanced models anytime and anywhere, such as augmented reality or real-time translation~\cite{hua2024interactivespeculativeplanningenhance,yao2024minicpm,chu2023mobilevlm}. {Lastly}, it drives research toward minimizing computational and memory demands to ensure efficient operation on resource-constrained hardware~\cite{wang2024cloud,xue2024powerinfer}.

However, deploying LLMs and MLLMs on mobile phones remains challenging. Firstly, the limited memory capacity of mobile phones restricts the deployment of large-parameter models. For example, a 4-bit quantized LLaMA 7B model requires approximately 4.5 GB of memory, which can impact system fluency due to high memory usage. Secondly, the limited computational power of mobile processors constrains inference speed. For instance, on the MediaTek Dimensity 9300 processor, a 4-bit quantized LLaMA 7B model generates around 10-15 tokens per second, limiting its suitability for real-time applications. Thirdly, mainstream MLLMs~\cite{chen2024far,liu2024llavanext} often use dynamic image resolution strategies to enhance high-resolution image understanding, leading to multiple inferences of ViT and excessive image tokens. This hinders image processing speed and affects overall latency for end-side deployment.

To address these challenges, we propose \textbf{BlueLM-V-3B}, an algorithm and system co-design approach that enables more efficient deployment of MLLMs on mobile devices. Specifically, we train a state-of-the-art MLLM with only 3B parameters and effectively deploy it on the NPU of smartphones. In terms of \textit{\textbf{algorithm}} design, we find that traditional dynamic resolution schemes~\cite{chen2024far,liu2024llavanext} lead to exaggerated image enlargement, resulting in longer image tokens and complicating mobile deployment. We propose a relaxed aspect ratio matching method, which effectively reduces the number of image tokens without sacrificing model accuracy. In terms of \textit{\textbf{system}} design, different from previous papers on end-side MLLM~\cite{yao2024minicpm,chu2023mobilevlm}, we incorporate a detailed system optimization for hardware-aware deployment. To accelerate image encoding, we design batched image encoding together with pipeline parallelism processing for the image patches generated by the dynamic resolution processor. To address the inefficiency of NPU when processing long input tokens, we adopt a token downsampling method and implement a chunked computing approach. We also enhance deployment efficiency by using mixed-precision deployment and carefully designing the overall inference framework.

Compared to previous efforts for efficient MLLM deployment on mobile phones~\cite{yao2024minicpm,chu2024mobilevlm,chu2023mobilevlm}, BlueLM-V-3B achieves higher model performance and features a more detailed algorithm and system co-design, paving the way for more powerful and efficient MLLMs optimized for mobile environments. The features of BlueLM-V-3B and the contributions of our work are summarized as follows:

\textbf{1) Algorithm and System Initiative}: We identify and address the excessive image enlargement issue in the dynamic resolution scheme used by classical MLLMs. Additionally, we implement a series of system designs and optimizations for hardware-aware deployment, resulting in more efficient inference of MLLMs on mobile devices.

\textbf{2) State-of-the-art MLLM Performance}: BlueLM-V-3B achieves SOTA performance (e.g., 66.1 on the OpenCompass benchmark) among models with similar parameter sizes, even surpassing a series of MLLMs with much more parameters (e.g., MiniCPM-V-2.6, InternVL2-8B).

\textbf{3) High Deployment Efficiency}: BlueLM-V-3B is highly efficient when deployed on mobile phones. Take the MediaTek Dimensity 9300 processor as an example, with a memory requirement of just 2.2GB, it can encode images with a resolution of 768$\times$1536 in approximately 2.1 seconds and achieves a token throughput speed of 24.4 token/s.

\section{Related Works}\label{related}

\subsection{Multimodal Large Language Models}
Large language models (LLMs)~\cite{brown2020language, touvron2023llama1, touvron2023llama2, anil2023palm} have demonstrated impressive success in tackling various complex tasks~\cite{wei2022emergent, trinh2024solving} in recent years. Building upon these advancements, multimodal large language models (MLLMs) incorporate visual inputs into LLMs~\cite{Achiam2023GPT4TR, liu2023llava, chen2024internvl, zhang2023llamaadapter} to handle multimodal scenarios. Various methods have been designed to integrate visual knowledge into language models, such as the linear projector approach~\cite{liu2023llava, chen2024far, wang2024qwen2}, the Q-Former approach~\cite{li2023blip}, and the perceiver resampler approach~\cite{alayrac2022flamingo, bai2023qwen,yao2024minicpm}. To further enhance MLLMs' capability to comprehend high-resolution images, a dynamic resolution scheme has recently been proposed~\cite{chen2024far,liu2023improvedllava,liu2024llavanext}. This scheme enables the model to adaptively process images at different resolutions while capturing more detailed information~\cite{huang2024mini}. However, when deploying on mobile devices, the dynamic resolution approach presents two challenges: firstly, an excessive number of image patches can significantly slow down the processing speed of the image encoder, and secondly, long sequences of image tokens can result in increased latency in the language model~\cite{lin2023vila}.

\begin{figure}[t]
    \centering
    \includegraphics[width=\linewidth]{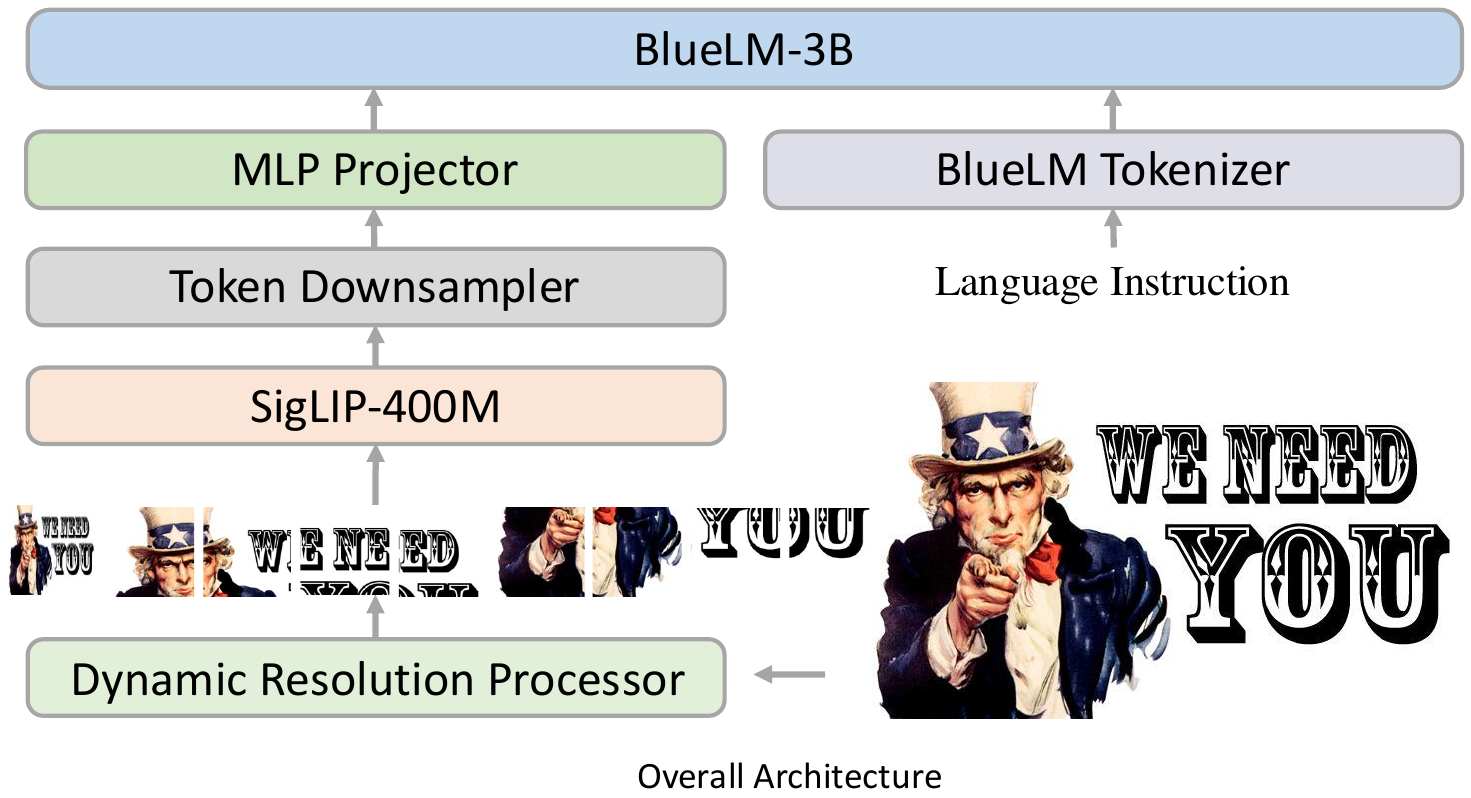}
    \vspace{-1.8em}
    \caption{\textbf{Model architecture of BlueLM-V-3B.} The architecture of BlueLM-V-3B follows the classical LLaVA approach. We integrate a dynamic resolution processing module (as in LLaVA-NeXT~\cite{liu2024llavanext} and InternVL 1.5~\cite{chen2024far}) to enhance model capabilities and apply token downsampling to reduce deployment complexity.}
    \label{fig:bluelm_v}
    \vspace{-1.2em}
\end{figure}

\begin{figure*}[t]
    \centering
    \includegraphics[width=\textwidth]{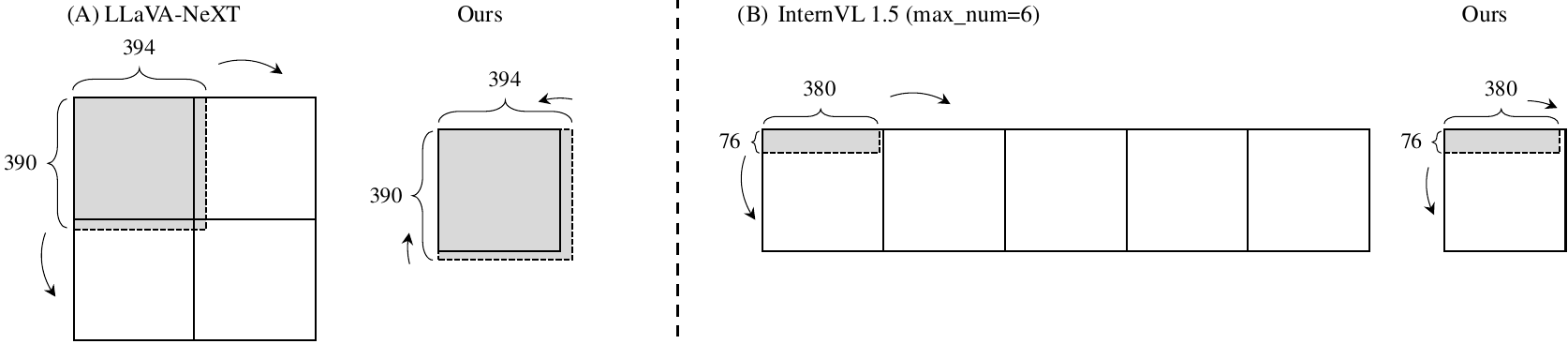}
    \vspace{-1.7em}
    \caption{\textbf{Existing methods overly enlarge images.} (A) For LLaVA-NeXT, an image with resolution 394$\times$390 selects a 2:2 aspect ratio and is resized and padded to 768$\times$768 (4$\times$ area enlargement). (B) For InternVL 1.5, an image with resolution 380$\times$76 chooses a 5:1 aspect ratio and is directly resized to 1920$\times$384 (25$\times$ area enlargement). BlueLM-V-3B, in contrast, selects a 1:1 aspect ratio for both resolutions, resulting in the minimum number of image tokens after ViT encoding, which can facilitate both model training and deployment.}
    \vspace{-0.5em}
    \label{fig:dynamic_question}
\end{figure*}

\subsection{On-device Large Language Models}
As application scenarios for large models continue to expand, small-scale large language models (SLMs) are now attracting increasing attention as consumers seek more cost-effective and efficient solutions~\cite{ashkboos2024computational}. A series of SLMs have emerged to meet these demands, including language models~\cite{hu2024minicpm,abdin2024phi,mehta2024openelm} and multimodal language models~\cite{yao2024minicpm,wang2024qwen2,chen2024far,luo2024mono,li2024llava}. The compact size of these models (2-3B parameters) enables deployment on user devices, such as personal computers and mobile phones. In addition to developing smaller LLMs and MLLMs with higher performance, recent studies from a system perspective have also introduced various methods for deploying these SLMs on user devices, such as personal computers~\cite{wei2024t}, and mobile phones~\cite{yao2024minicpm,li2024transformer}. Our proposed BlueLM-V-3B is an algorithm and system co-design approach that not only achieves state-of-the-art model capability but also enables efficient deployment of MLLMs on mobile devices.

\section{BlueLM-V-3B}\label{method}

We provide a detailed introduction of BlueLM-V-3B in this section, emphasizing its model architecture and highlighting the algorithm and system co-design that optimizes efficiency during both the training and deployment stages.

\subsection{Basic Network Components}
\textbf{Model Architecture}: Our architecture is modified from the classical LLaVA approach~\cite{liu2023llava}, as it has been shown effective in prior works such as InternVL 1.5~\cite{chen2024far} and LLaVA-NeXT~\cite{liu2024llavanext}. The overall architecture is shown in Fig.~\ref{fig:bluelm_v}. It is composed of the following components. \textbf{\textit{Image Encoder}}: To process multimodal (image and language) inputs, we utilize the SigLIP~\cite{zhai2023sigmoid} ViT for 384$\times$384 input images, as described in~\cite{lin2023vila,yao2024minicpm}, which has 400M parameters. \textbf{\textit{MLP Projector}}: A 2-layer MLP is employed to map the space of image tokens to LLM tokens. \textbf{\textit{LLM}}: We employ an in-house 2.7B BlueLM model as the core language model to design BlueLM-V-3B. To further enhance the model's ability to understand high-resolution images, a \textbf{\textit{Dynamic Resolution Processor}} module is integrated. We reflect on the exaggerated image enlargement issue observed in InternVL 1.5~\cite{chen2024far} and LLaVA-NeXT~\cite{liu2024llavanext}, then introduce a new approach that improves both training and deployment efficiency. Given the limited performance of NPUs in handling long tokens, we leverage a \textbf{\textit{Token Downsampler}} module to reduce deployment complexity, which will be introduced in Sec.~\ref{sec:dynamic_resolution} and Sec.~\ref{sec:token_downsample} respectively.\\

\hspace{-1.2em}\textbf{Training and Inference}: During training, the Image Encoder receives input images processed by the Dynamic Resolution Processor (for multiple input images, we simply concatenate them). The output features are passed through the Token Downsampler and MLP Projector to produce the corresponding image tokens. These image tokens are then concatenated with the language instruction tokens provided by the user. The resulting token sequences are used for model training. For inference, image and text tokens are similarly obtained (with user instructions in the audio format being converted to text first), and the model generates subsequent tokens in an autoregressive manner.

\subsection{Dynamic Image Resolution}~\label{sec:dynamic_resolution}
\hspace{-8px}The original ViT of LLaVA~\cite{liu2023llava} directly resizes the input images to a fixed resolution (e.g., 336$\times$336 or 384$\times$384), which is not well-suited for high-resolution images. To address this issue, BlueLM-V-3B adopts a dynamic image resolution design, which has been proven effective in InternVL 1.5~\cite{chen2024far} and LLaVA-NeXT~\cite{liu2024llavanext}. We observe exaggerated image enlargement issues in these two methods and make improvements for better training and easier deployment. Additionally, we design a batched image patch encoding with pipeline parallelism to further enhance the efficiency of both training and inference.\\

\hspace{-1.2em}\textbf{Exaggerated Image Enlargement}: LLaVA-NeXT~\cite{liu2024llavanext} and InternVL 1.5~\cite{chen2024far} both propose dynamic image resolution approaches to tackle high-resolution images. For the SigLIP encoder, both approaches use 384$\times$384 as the base resolution (1:1), then select an appropriate resolution aspect ratio ($m$:$n$) to resize (and pad) the original image to a size of 384$m\times$384$n$. The image is subsequently divided into patches of 384$\times$384. LLaVA-NeXT tends to select aspect ratios that result in a larger image than the original one but with a smaller total area, while InternVL 1.5 opts for the ones that match the original image's width-to-height ratio.

We use some examples in Fig.~\ref{fig:dynamic_question} to demonstrate the exaggerated image enlargement by the two methods. For LLaVA-NeXT in Fig.~\ref{fig:dynamic_question}A, given an image with 394$\times$390 resolution, it will choose a ratio of 2:2, then resize and pad the original image to 768$\times$768 (4$\times$ area enlargement). For InternVL 1.5 in Fig.~\ref{fig:dynamic_question}B, given an image with 380$\times$76 resolution (setting max\_num=6, i.e., $m\times n\leq 6$), it will choose a ratio of 5:1, then directly resize the original image to 1920$\times$384 (25$\times$ enlargement)\footnote{The recently proposed adaptive gridding method in Ferret-UI 2~\cite{li2024ferret} shares a similar issue with InternVL 1.5, as strictly preserving the aspect ratio (e.g., choosing 5:1 instead of 1:1 for a \(380 \times 76\) image) results in \(\Delta_\text{aspect} = 0\), according to the pseudocode provided in the paper.}. The enlargement does not necessarily alter image information, but increases deployment difficulty on mobile devices, as higher resolutions lead to more image patches and thus longer image tokens. Therefore, we redesign the dynamic resolution approach, as will be discussed in the following paragraph.\\

\hspace{-1.2em}\textbf{Relaxed Aspect Ratio Matching}: We propose a \textbf{relaxed} aspect ratio matching method based on LLaVA-NeXT to mitigate the exaggerated image enlargement. LLaVA-NeXT defines effective resolution $R_e$ and wasted resolution $R_w$. For aspect ratios ($m$:$n$) that lead to image resolution 384$m\times$384$n$ smaller than the original image in either dimension, \( R_e \) is defined as the maximum area of the image scaled proportionally to fit within the dimensions 384$m\times$384$n$. Otherwise, $R_e$ is set to the area of the original image. After getting $R_e$, $R_w$ is calculated by 384$m\times$384$n$$-R_e$. LLaVA-NeXT explores feasible aspect ratios and searches for the largest $R_e$ while minimizing $R_w$. Generally, larger aspect ratio options are available because \(m\) and \(n\) are always set to bigger values to accommodate high-resolution images (e.g., from 1:1 to 6:6 in LLaVA-OneVision~\cite{li2024llava}). In this context, LLaVA-NeXT often leads to image enlargement by selecting an aspect ratio that offers a higher resolution than the original image, resulting in the cases shown in Fig.~\ref{fig:dynamic_question}A. Considering that larger resolutions (4$\times$) do not necessarily provide additional information but rather increase training and deployment complexity, downscaling the image is a more appropriate choice.

We propose a relaxed aspect ratio matching method by leveraging a threshold to prevent the trend of always selecting larger resolutions. To be specific, we add a parameter 
$\alpha$\footnote{$\alpha=0.1$ in our implementation.} such that when:
\begin{equation}
R_e - R_{e, \text{max}} > \alpha \cdot R_{e, \text{max}},
\end{equation}
or
\begin{equation}
(R_{e, \text{max}} - R_e) < \alpha \cdot R_{e, \text{max}}\; \text{and} \; R_w < R_{w, \text{min}},
\end{equation}
we then update
\begin{equation}
R_{e, \text{max}} \leftarrow R_e, \quad R_{w, \text{min}} \leftarrow R_w,
\end{equation}
and record the according aspect ratio. In our design, smaller $R_e$ with smaller $R_w$ will have a chance to be chosen. To further increase the likelihood of selecting a smaller $R_w$, we enumerate the candidate aspect ratios in descending order, e.g., from 3:3 to 1:1 (assuming that a smaller aspect ratio leads to a smaller total area, thus a smaller $R_w$). Our relaxed aspect ratio matching method allows for more flexible handling of dynamic resolution. As shown in Fig.~\ref{fig:dynamic_question}, when facing these two extreme cases, our solution can still select the appropriate aspect ratio (1:1). 
The pseudocode of the relaxed aspect ratio matching method is shown in Alg.~\ref{alg:pseudocode}.\\

\begin{figure}[t]
\vspace{-0.5em}
    \centering
    \includegraphics[width=\linewidth]{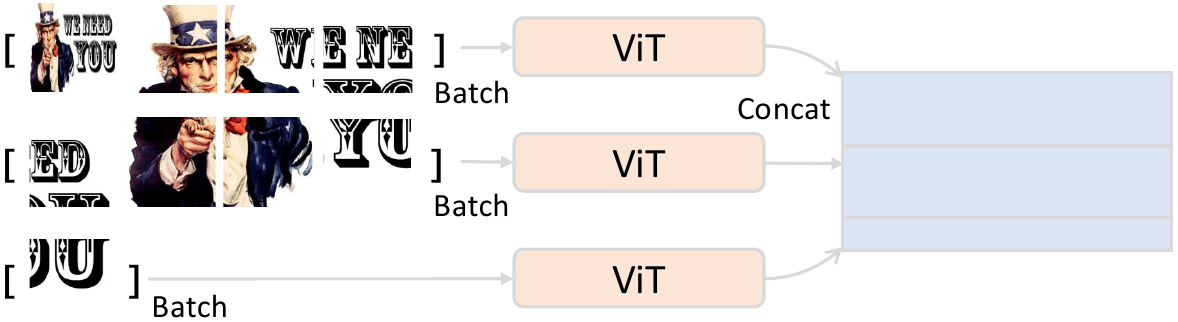}
    \vspace{-1.7em}
    \caption{\textbf{Batched image encoding on NPU.} We design a parallel processing scheme for image patches on the NPU. The figure illustrates the case of 4 patches being processed in parallel.}
    \label{fig:vit_paral}
    \vspace{-1.7em}
\end{figure}

\begin{figure*}[t]
    \vspace{-1em}
    \centering
    \includegraphics[width=0.9\linewidth]{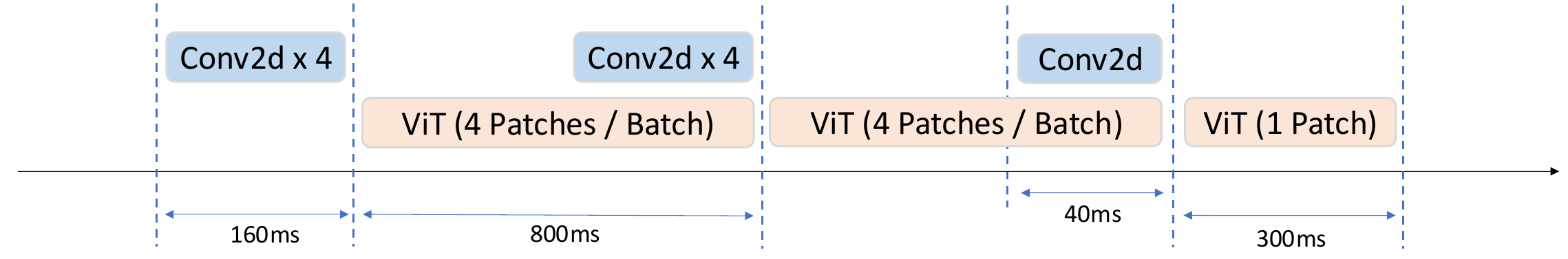}
    \vspace{-0.5em}
    \caption{\textbf{Pipeline parallelism in image encoding.} We design a pipeline parallelism scheme for image encoding. The Conv2D layer in the vision embedding module of SigLIP (on the CPU) and the vision transformer blocks (on the NPU) for different image patches run parallel to improve inference speed. This image illustrates the pipeline parallelism scheme combined with batched image patch encoding.}
    \label{fig:parallel}
    \vspace{-1.2em}
\end{figure*}

\hspace{-1.2em}\textbf{Batched Image Patch Encoding}: We also propose system optimization to achieve more efficient hardware-aware training and deployment. After the dynamic resolution processing, an image is divided into several local patches, together with a thumbnail image (global patch). In the training process, we batch the image patches before inputting them into the ViT, leveraging GPU parallelism to accelerate the process and achieving a 10\% speedup. For inference, we adopt a similar parallel strategy to exploit the NPU's computing capabilities. Unlike high-level languages (e.g., Python), hardware acceleration design requires low-level control over computing resources, such as memory layout and computational optimizations based on register size. Due to the NPU's limited computational power, all patches cannot be effectively processed simultaneously; instead, we handle a fixed batch size at a time. Fig.~\ref{fig:vit_paral} illustrates the concurrent processing of 4 patches for a 2:4 aspect ratio (the ratio we use to process mobile phone screens) following a $4+4+1$ approach. This concurrent patch processing notably reduces overall processing time.\\

\hspace{-1.2em}\textbf{Pipeline Parallelism in Image Patch Encoding}: During the model inference process, we implement a pipeline parallelism scheme to optimize image patch encoding. Specifically, for different patches extracted from a single image, we design a parallel pipeline for the Conv2D layer in SigLIP's vision embedding module (on the CPU) and the vision transformer blocks (on the NPU). We show the encoding pipeline on MediaTek Dimensity 9300 processor with a 2:4 aspect ratio in Fig.~\ref{fig:parallel}. This approach helps to conceal the execution latency of the Conv2D operation.\\

\subsection{Token Downsampler}~\label{sec:token_downsample}
\hspace{-0.6em}Although we have designed a relaxed aspect ratio matching method to mitigate the exaggerated image enlargement, the dynamic image resolution strategy still results in a significant number of image tokens, posing challenges for the deployment on mobile phone processors and potentially exceeding the maximum context length of the language model. For instance, with an image that selects a 2:4 aspect ratio (the resolution we use for processing mobile phone screens), we obtain a total of 9 image patches (calculated as $2\times4+1$). This results in $729\times9=6561$ image tokens from SigLIP after the dynamic resolution processor, making it too long to deploy on the NPU.\\

\hspace{-1.2em}\textbf{Basic Practice}: To reduce the excessive number of image tokens, we apply the downsampler block proposed in VILA~\cite{lin2023vila}. Specifically, it concatenates every $2\times2$ tokens into a single token and then employs a linear layer to fuse the information. This effectively reduces the number of image tokens generated by SigLIP from 729 to 196, resulting in a total of $196\times9=1764$ image tokens for the 2:4 aspect ratio setting. However, the approximately 2k length of the image tokens, when combined with the user instruction, still poses challenges for deployment on the NPU, as will be discussed in the following paragraph.\\

\hspace{-1.2em}\textbf{Chunked Computing of Input Tokens}: During the inference process of 
a LLM, to accelerate the computation of input tokens, traditional GPUs frequently employ parallel computing techniques to process all input tokens simultaneously. However, the excessive length of these tokens (due to the extended length of image tokens or the contextual information involved), combined with the limited computational capacity of NPUs, renders the parallel processing of all input tokens inefficient. Conversely, sequential processing of individual tokens (t1) is also suboptimal. Consequently, we implement a chunking strategy on mobile devices, processing 128 input tokens in parallel (t128) per iteration, and then combining the results. This approach strikes a balance between parallel processing and the computational resources available on the NPU.
\hspace{-1.2em}\subsection{Model Quantization and Overall Framework}\label{sec:deploy}
With the above design and optimization, we deploy the BlueLM-V-3B model on the MediaTek Dimensity 9300 processor. We hope to take full advantage of the device's capabilities, offering a powerful yet efficient solution for running the model in a mobile environment.\\

\hspace{-1.2em}\textbf{Mixed Parameter Precision}: We apply mixed-precision quantization to further reduce memory usage and improve inference speed. We employ INT8 precision for the ViT and the MLP projector weights, and INT4 precision for the LLM weights. This combination strikes a balance between computational efficiency and model accuracy. However, we find that the activation values are more sensitive to quantization; therefore, we maintain INT16 precision for LLM activation and FP16 for ViT and projector activation to ensure the model's performance remains robust. During inference, we store the KV cache in INT8 precision.\\

\begin{figure}[t]
    \centering
    \includegraphics[width=\linewidth]{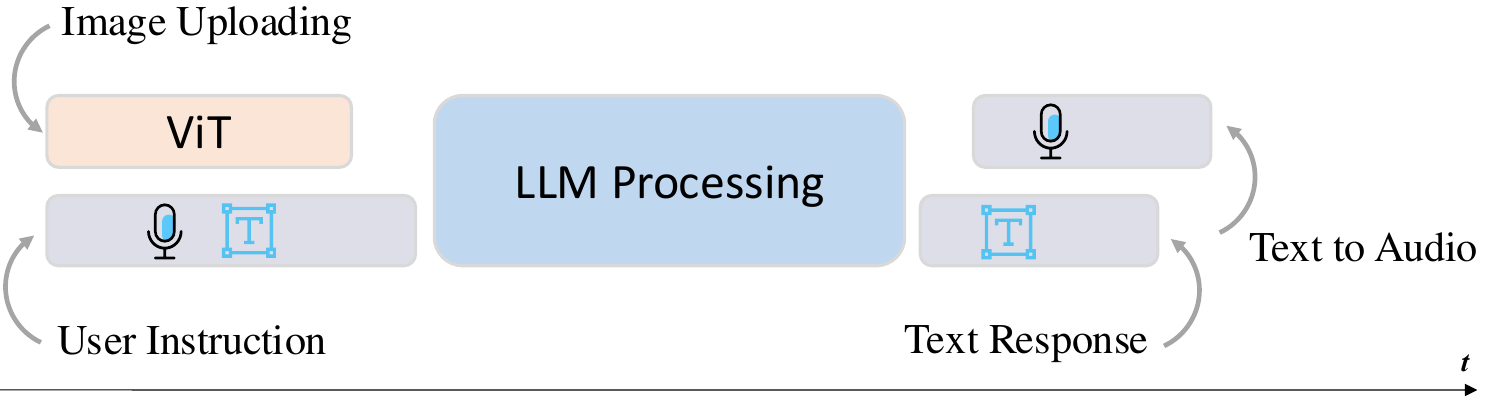}
    \vspace{-1.5em}
    \caption{\textbf{Overall framework of deploying BlueLM-V-3B.} We decouple ViT image processing from user instruction (text or audio) handling to enhance overall efficiency. The text responses by LLM can be further converted on the fly to audio responses.}
    \label{fig:decoupling_patch}
    \vspace{-1.5em}
\end{figure}

\hspace{-1.2em}\textbf{Decoupling Image Encoding and Instruction Processing}: To improve overall efficiency during deployment, we decouple image processing from user input handling. At model initialization, we load both the ViT and LLM models simultaneously. Users begin by uploading an image, and since the MLLM is deployed locally, this upload takes virtually no time. The ViT starts processing the image immediately after the upload is complete. Meanwhile, users can input their instructions simultaneously; for instructions in audio format, we convert them to text first. Once the image processing is finished, the user's commands are submitted to the LLM for response generation, and the ViT can be freed from memory. This parallel process, illustrated in Fig.~\ref{fig:decoupling_patch}, reduces the waiting time for the first token generation, improves overall responsiveness, and limits the peak memory usage of BlueLM-V-3B to 2.2GB.

\section{Training Recipe}\label{sec:train_recipe}
In this section, we detail the training process and training data for BlueLM-V-3B.

\subsection{Training Process}

We begin with the BlueLM-3B language model and train the model in two stages. In the first stage, We pre-train the MLP projection layer while keeping the ViT and LLM frozen. In the second stage, we fully fine-tune the model with a large set of image-text pairs.

\subsection{Training Data}

\textbf{Pre-training Stage:} The pre-training stage aims to equip the model with basic multimodal capabilities. In this stage, we utilize open-source datasets, creating a comprehensive pre-training dataset composed of 2.5 million image-caption pairs drawn from LLaVA 558k~\cite{liu2023improvedllava}, ShareGPT4V 1200k~\cite{chen2023sharegpt4v}, and ALLaVA 708k~\cite{chen2024allava}.\\

\hspace{-1.2em}\textbf{Fine-tuning Stage:} During the fine-tuning process, we meticulously construct a dataset containing 645 million image-text pairs, including both open-source and in-house datasets. This dataset covers a variety of downstream tasks and diverse dataset types, such as captioning, VQA, OCR, and pure text. Tab.~\ref{tab:data_ratio} summarizes the distribution of data types, along with the proportions of public and in-house data in our fine-tuning dataset.

To better illustrate the data we use, we present the open-source datasets utilized at this stage in Tab.~\ref{tab:data}. In addition to these open-source datasets, we also incorporate in-house data to enhance the model's capabilities. We crawl a significant amount of pure text data and image-text pairs from various websites. For different data categories, we also manually create a large number of image-text pairs to enrich the diversity of the training data.

For PDF documents, we utilize the \texttt{PyMuPDF}\footnote{https://github.com/pymupdf/PyMuPDF} library to convert them into image-text pairs. For formula data, we use \texttt{Matplotlib}\footnote{https://github.com/matplotlib/matplotlib} to render them into necessary representations. For table contents, we convert the data to Markdown format and render it with the \texttt{IMGKit}\footnote{https://github.com/csquared/IMGKit} library. For problem-solving data, we similarly convert the text into Markdown format and render it using the \texttt{IMGKit} library. Additionally, we manually render a substantial amount of multilingual OCR data by converting texts in various languages into image-text pairs, which helps to enhance the model's multilingual understanding and capabilities.

In addition to image rendering, we also utilize GPT-4o~\cite{Achiam2023GPT4TR} and Gemini Pro~\cite{team2024gemini} to create and revise image captions and question-answering pairs. The combination of open-source and proprietary data significantly enhances the model's capabilities, allowing it to learn from a diverse range of examples and improve its performance across various tasks and modalities.

\begin{table}[t]\small
\vspace{-0.5em}
\resizebox{\columnwidth}{!}{%
\begin{tabular}{lccc}
\midrule
\multicolumn{1}{l}{\textbf{Type}} & \textbf{Public (M)} & \textbf{In-House (M)} & \textbf{In-House / Public} \\ \midrule
\textbf{Pure Text} & 2.2 & 64.7 & 29.4 \\ 
\textbf{Caption} & 10.0 & 306.3 & 30.6 \\ 
\textbf{VQA} & 20.3 & 44.4 & 2.2 \\ 
\textbf{OCR} & 23.3 & 173.9 & 7.5 \\ \midrule
\textbf{Total} & \textbf{55.8} & \textbf{589.3} & \textbf{10.6} \\ \midrule
\end{tabular}%
}
\vspace{-0.5em}
\caption{\textbf{Detailed statistics of the fine-tuning dataset.} Summary of dataset types, counts (in millions), and in-house/public ratios for each category used in fine-tuning.}
\label{tab:data_ratio}
\vspace{-1.5em}
\end{table}

\section{Experiments}\label{experiment}

In this section, we conduct a series of experiments to validate the effectiveness of our proposed approaches and to demonstrate the capabilities of BlueLM-V-3B in terms of benchmark accuracy and deployment efficiency.

\begin{table*}[t]\Large
\resizebox{\textwidth}{!}{%
\renewcommand\arraystretch{0.95}
\begin{tabular}{llclcccccc}
\midrule
\textbf{Language Model} &\textbf{Vision Model} & \textbf{Params} & \textbf{Method} & \textbf{VQAv2}$_\text{val}$ & \textbf{TextVQA}$_\text{val}$ & \textbf{DocVQA}$_\text{val}$ & \textbf{OCRBench} & \textbf{ChartQA}$_\text{test}$ \\ \midrule
\multirow{3}{*}{\centering \textbf{MiniCPM-2B}~\cite{hu2024minicpm}}&\multirow{3}{*}{\centering \textbf{SigLIP-400M}~\cite{zhai2023sigmoid}}  & \multirow{3}{*}{\centering 3B} & InternVL 1.5 & 70.5 & 46.9 & 26.2 & 327 & 15.7 \\ 
 & & & LLaVA-NeXT & 70.1 & 44.2  & 24.3 & 324 & 14.8 \\ \cmidrule{4-9} 
 & & & Ours & \textbf{71.8} & \textbf{49.4} & \textbf{27.3} & \textbf{343} & \textbf{16.9} \\ \midrule
\multirow{4}{*}{\centering \textbf{BlueLM-3B}} &\multirow{4}{*}{\centering \textbf{SigLIP-400M}~\cite{zhai2023sigmoid}}  & \multirow{4}{*}{\centering 3B} & InternVL 1.5 & 78.3 & 52.7 & 28.7 & 338 & 16.8 \\ 
 & & & LLaVA-NeXT & 77.7 & 51.4 & 29.6 & 351 & 16.4 \\ \cmidrule{4-9} 
 & & & Ours & \textbf{79.5} & \textbf{56.2} & \textbf{31.3} & \textbf{360} & \textbf{17.5} \\ 
 \rowcolor{gray!20} \cellcolor{white}& \cellcolor{white} & \cellcolor{white} & Ours (fully-trained) & {82.7} & {78.4} & {86.6} & {829} & {80.4} \\ \midrule

\end{tabular}%
}
\vspace{-0.5em}
\caption{\textbf{Comparison results of different dynamic resolution methods.} We compare the performance of models trained using different dynamic resolution methods. We use the LLaVA~\cite{liu2023improvedllava} 558k dataset for pre-training, and the LLaVA 665k dataset for fine-tuning. To better demonstrate our improvements, we conduct experiments on both our in-house BlueLM-3B language model and the open-sourced MiniCPM-2B language model, which have similar parameter counts (2.7B). Our dynamic image processing method achieves the best performance. $^\dagger$We also provide the results of the fully trained BlueLM-V-3B model for reference.}
\label{tab:minicpm}
\end{table*}

\begin{table*}[t]\Large
\renewcommand\arraystretch{1.1}
\resizebox{\textwidth}{!}{%
\begin{tabular}{lcccccccccc}
\midrule
\textbf{Model} & \textbf{Params} & \textbf{Avg.} & \textbf{MMBench} & \textbf{MMStar} & \textbf{MMMU} & \textbf{MathVista} & \textbf{HallusionBench} & \textbf{AI2D} & \textbf{OCRBench} & \textbf{MMVet} \\ \midrule
\textbf{Qwen2-VL}~\cite{wang2024qwen2} & 8B & \textbf{67} & 81 & 60.7 & \textbf{53.7} & 61.4 & \textbf{50.4} & 83 & 843 & \textbf{61.8} \\ 
\textbf{MiniCPM-V-2.6}~\cite{yao2024minicpm} & 8B & 65.2 & 78 & 57.5 & 49.8 & 60.6 & 48.1 & 82.1 & \textbf{852} & 60 \\ 
\textbf{InternVL2}~\cite{chen2024far} & 8B & 64.1 & 79.4 & 61.5 & 51.2 & 58.3 & 45 & 83.6 & 794 & 54.3 \\ 
\textbf{POINTS-Qwen2.5}~\cite{liu2024points} & 8.3B & 62.5 & 78 & 60.9 & 51.4 & \textbf{63} & 45.6 & 81.2 & 717 & 47.9 \\  \midrule
\rowcolor{gray!20} \textbf{BlueLM-V} (Ours)  & \textbf{3B} & 66.1 & \textbf{82.7} & \textbf{62.3} & 45.1 & 60.8 & 48 & \textbf{85.3} & 829 & \textbf{61.8} \\ \midrule
\end{tabular}%
}
\vspace{-0.5em}
\caption{\textbf{OpenCompass benchmark.} Comparison results on the OpenCompass benchmark for models with parameter sizes less than or equal to \textbf{10B}. BlueLM-V-3B achieves state-of-the-art performance on 4 out of 8 tasks, with an average performance ranking of second.}
\vspace{-0.5em}
\label{exp:opencompass}
\end{table*}

\begin{table}[t]\Large
\renewcommand\arraystretch{1.2}
\resizebox{\columnwidth}{!}{%
\begin{tabular}{lcccccc}
\midrule
\textbf{Model} & \textbf{Params} & \textbf{TextVQA}$_\text{val}$ & \textbf{DocVQA}$_\text{test}$  & \textbf{MTVQA} \\ \midrule
\textbf{Phi-3-Vision}~\cite{abdin2024phi3} & 4.2B & 72.4 & 84.6 & 13.9 \\ 
\textbf{MiniCPM-V-2}~\cite{yao2024minicpm} & 2.8B & 73.2 & 71.9 & 9.3 \\ 
\textbf{InternVL2}~\cite{chen2024far} & 4B & 74.7 & 89.2 & 15.5 \\ 
\textbf{Qwen2-VL}~\cite{wang2024qwen2} & \textbf{2B} & \textbf{79.9} & \textbf{90.1} &  20.7 \\ \midrule
\rowcolor{gray!20} \textbf{BlueLM-V} (Ours) & 3B & 78.4 & 87.8 & \textbf{32.7} \\ \midrule
\end{tabular}%
}
\vspace{-0.5em}
\caption{\textbf{Text-centric/OCR benchmarks.} Comparison on text-centric/OCR benchmarks shows that BlueLM-V-3B achieves performance comparable to SOTA MLLMs with similar parameter sizes, while significantly enhancing multilingual capability. $^\dagger$We evaluate TextVQA and MTVQA on VLMEvalKit~\cite{duan2024vlmevalkit} for a fair comparison. OCRBench has been included in OpenCompass.}
\vspace{-0.5em}
\label{tab:ocr}
\end{table}

\begin{table*}[t]
\resizebox{\textwidth}{!}{%
\begin{tabular}{lcccccc}
\midrule
\textbf{Model   Name} & \textbf{Params} & \textbf{Processor} & \textbf{Solution} & \textbf{Image Processing} & \textbf{LLM Prefilling} & \textbf{Throughput} \\ \midrule
MiniCPM-V   2.5~\cite{yao2024minicpm} & 8B  & MediaTek Dimensity 9300 & CPU (llama.cpp) \frownie{} & 4.0s & 13.9s & 4.9 token/s \\ \midrule
\rowcolor{gray!20} BlueLM-V-3B (Ours) & 3B & MediaTek Dimensity 9300 & NPU \smiley{} & \textbf{2.53s} (0.47+2.06) & \textbf{2.7s} & \textbf{24.4} token/s \\ \midrule
\end{tabular}%
}
\vspace{-0.7em}
\caption{\textbf{Deployment efficiency comparison with MiniCPM-V.} MiniCPM-V deploys an 8B model on the CPU, leading to longer image processing latency, LLM prefilling latency, and lower throughput. $^\dagger$MiniCPM-V calculates encoding latency by including both model loading time and encoding time. In our setting, we need 0.47s to simultaneously load the ViT and LLM once during system initialization.}
\label{tab:deploy_minicpm}
\vspace{-0.5em}
\end{table*}

\subsection{Relaxed Aspect Ratio Matching}

In BlueLM-V-3B, we alleviate the issue of ineffective image upscaling present in InternVL 1.5 and LLaVA-NeXT by applying a relaxed aspect ratio matching approach. In this section, we validate the enhancements by measuring the improvements in both deployment efficiency and benchmark accuracy, with both open-source and in-house models.\\

\hspace{-1.2em}\textbf{Deployment Efficiency:} We conduct a statistical analysis on the multimodal LLaVA~\cite{liu2023improvedllava} 665k training dataset. We omit the 41k text-only data and compare the aspect ratio selected by our approach with those chosen by the LLaVA-NeXT~\cite{liu2024llavanext} and InternVL 1.5~\cite{chen2024far} approaches. A larger aspect ratio selection results in higher image resolution, which leads to a longer image token sequence during deployment. We provide 9 candidate aspect ratios (from 1:1 to 3:3) for LLaVA-NeXT and our approach. For InternVL 1.5, we fix max\_num=9 and propose the initial candidate aspect ratios using the method described in the original paper. This will result in the same maximum area (9$\times$384$\times$384) for all approaches, and ensure a fair comparison.

Compared with LLaVA-NeXT, in \textbf{29k} cases, our method selects a smaller aspect ratio. Concerning InternVL 1.5, we select smaller aspect ratios in \textbf{523k} cases and larger aspect ratios in \textbf{25k} cases. This leads to a significant improvement in efficiency during the inference on NPU.\\

\hspace{-1.2em}\textbf{Benchmark Accuracy:} To further evaluate the performance impact of reducing image tokens, we train on the LLaVA 1.5~\cite{liu2023improvedllava} dataset with 3 dynamic image resolution approaches. We use the 558k data for pre-training and the 665k data for fine-tuning. Due to the slower learning capacity of the 3B model compared to the 7B/13B models, we train each stage for two epochs.

We conduct experiments on both our in-house BlueLM-3B model and the open-sourced MiniCPM-2B language model~\cite{hu2024minicpm}. The MiniCPM-2B model has 2.7B parameters, matching the parameter count of BlueLM-3B. The introduction of dynamic resolution is primarily aimed at handling high-resolution images, especially in OCR tasks. For evaluation, we use VQAv2~\cite{goyal2017making} for general vision question answering. For OCR tasks, we include TextVQA~\cite{singh2019towards}, DocVQA~\cite{mathew2021docvqa}, and OCRBench~\cite{liu2023hidden}. Additionally, we evaluate on ChartQA~\cite{masry2022chartqa} dataset for multimodal mathematical reasoning. Comparison results are shown in Tab.~\ref{tab:minicpm}. As can be seen, our proposed relaxed aspect ratio matching approach not only reduces the number of image tokens but also significantly improves accuracy. This indicates that simply enlarging images does not always lead to performance gains, as in LLaVA-NeXT and InternVL 1.5. The results demonstrate the effectiveness of our approach in both training outcomes and deployment efficiency.

\subsection{Accuracies on Different Benchmarks}

After extensive fine-tuning, we evaluate the performance of BlueLM-V-3B on various mainstream benchmarks.\\

\hspace{-1.2em}\textbf{OpenCompass Benchmark:} To evaluate the overall performance of BlueLM-V-3B, we use the OpenCompass benchmark~\cite{2023opencompass}, including MMbench~\cite{liu2025mmbench}, MMStar~\cite{chen2024we}, MMMU~\cite{yue2024mmmu}, MathVista~\cite{lu2023mathvista}, HallusionBench~\cite{guan2023hallusionbench}, AI2D~\cite{kembhavi2016diagram}, OCRBench~\cite{liu2023ocrbench}, and MM-Vet~\cite{yu2023mm}. BlueLM-V-3B demonstrates significantly improved performance over models with a similar number of parameters ($\leq$ 4B), as illustrated in Fig.~\ref{fig:radar_fig}. We also compare BlueLM-V-3B to models with up to 10B parameters, with detailed results presented in Tab.~\ref{exp:opencompass}. BlueLM-V-3B achieves SOTA performance on 4 out of 8 tasks, with an average performance ranking of second. With only 3B parameters, BlueLM-V-3B outperforms a series of MLLMs with significantly more parameters, such as InternVL2-8B~\cite{chen2024far}, MiniCPM-V-2.6 (8B)~\cite{yao2024minicpm}. This demonstrates that MLLMs with a smaller number of parameters can still achieve strong capacity.\\

\begin{figure}[t]
    \centering
    \includegraphics[width=\linewidth]{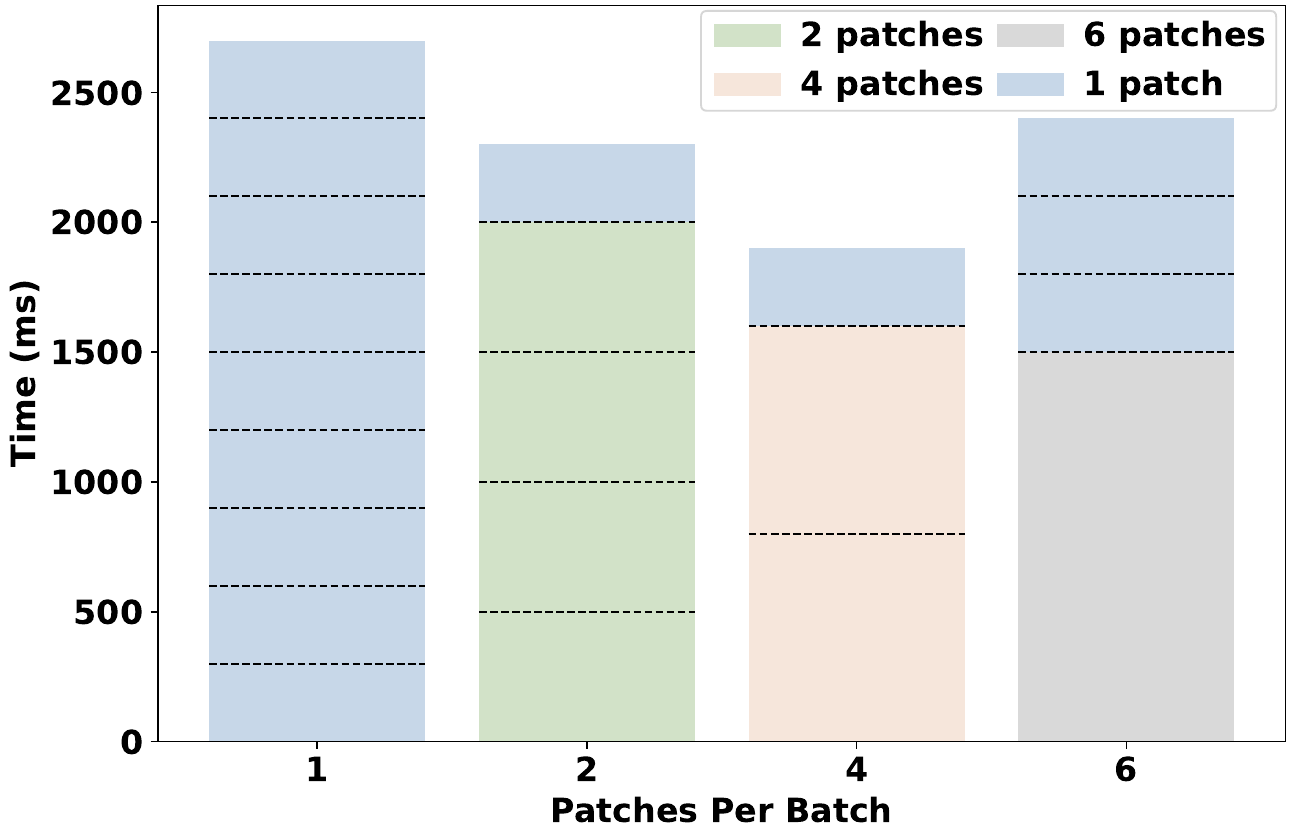}
    \vspace{-1.1em}
    \caption{\textbf{ViT inference time for 2:4 resolution aspect ratio.} We experiment with 1, 2, 4, and 6 image patches per batch on the NPU, using a 2:4 resolution aspect ratio (comprising one global patch and 8 local patches). Overall, processing 4 patches per batch delivers the fastest performance.}
    \vspace{-1.4em}
    \label{fig:vit_patch}
\end{figure}

\hspace{-1.2em}\textbf{Text-centric/OCR Capacity:} We compare BlueLM-V-3B with popular MLLMs of similar parameter sizes on text-centric/OCR benchmarks, including TextVQA~\cite{singh2019towards} and DocVQA~\cite{mathew2021docvqa} (OCRBench~\cite{liu2023ocrbench} has been included in OpenCompass). For MLLMs deployed on mobile devices for everyday use, especially in overseas markets or among foreign users, multilingual capabilities are essential. We also evaluate the multilingual multimodal capabilities using the MTVQA~\cite{tang2024mtvqa} dataset. The comparison results are shown in Tab.~\ref{tab:ocr}. BlueLM-V-3B achieves performance comparable to state-of-the-art MLLMs with similar parameter sizes, while significantly enhancing multilingual capability.

\subsection{Deployment Efficiency Evaluation}

In this section, we discuss the deployment efficiency of our proposed algorithm and system co-design in BlueLM-V-3B. We deploy BlueLM-V-3B on vivo X100 mobile phone with MediaTek Dimensity 9300 processor as an example.\\

\hspace{-1.2em}\textbf{Batched Image Patch Encoding on NPU:} We design a parallel processing method for encoding image patches generated by the dynamic resolution processor. Fig.~\ref{fig:vit_patch} shows the ViT inference latency when processing an image with a 2:4 aspect ratio (9 patches in total). As illustrated, parallelizing the inference with an increasing number of patches initially reduces latency, but after a certain point, latency begins to increase again. The fastest encoding speed is achieved with 4 patches processed in parallel.\\

\begin{figure}
    \centering
    \vspace{-0.3em}
    \includegraphics[width=\linewidth]{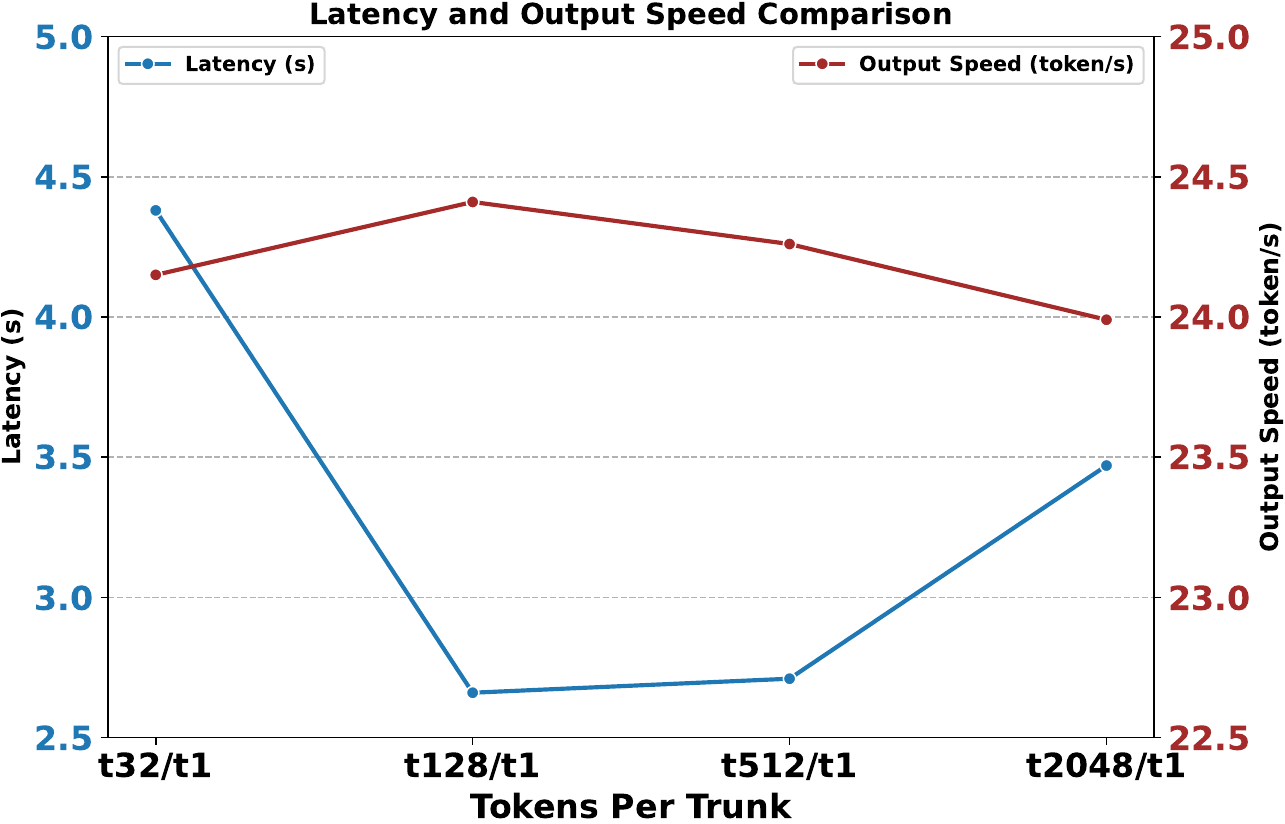}
    \vspace{-1.2em}
    \caption{\textbf{Latency and output speed comparison.} We compare the latency and output generation speed with processing different numbers of input tokens in parallel. t\{$x$\}/t1 implies processing $x$ input tokens in parallel. The output token is fixed to 1 per trunk as the LLM can only generate one token for each forward process.}
    \label{fig:latency_speed}
    \vspace{-1em}
\end{figure}

\hspace{-1.2em}\textbf{Pipeline Parallelism in Image Patch Encoding:} We design a pipeline parallelism scheme for the Conv2D layer and the vision transformer block in the SigLIP model. The processing pipeline for an image with a 2:4 aspect ratio is shown in Fig.~\ref{fig:parallel}, where we need 1.9 + 0.16 = 2.06s to encode the image and can hide a total latency of 200ms.\\

\hspace{-1.2em}\textbf{Chunked Computing of Input Tokens:} We implement a chunking strategy on the NPU to process 128 input tokens in parallel per iteration (t128), balancing parallel processing with NPU performance. Here, we present the LLM prefilling latency when processing different numbers of input tokens in parallel: t32, t128, t512, and t2048. For additional information, we also report the speed of output token generation, where only the t1 case is shown, as the LLM processes one token at a time for output. We fix the input token length to 2048 and set the KV cache length to 2048. As shown in Fig.~\ref{fig:latency_speed}, t128/t1 achieves the lowest latency and the fastest generation speed. Theoretically, the output token speed for t1 should remain consistent; however, we observe slight variations due to possible fluctuations in the testing environment.\\

\hspace{-1.2em}\textbf{Compasison with MiniCPM-V:} The MiniCPM-V paper~\cite{yao2024minicpm} only reports the deployment statistics for the 8B MiniCPM-V 2.5 model, which is deployed on the CPU of the MediaTek Dimensity 9300 using \texttt{llama.cpp}\footnote{https://github.com/ggerganov/llama.cpp}. However, the paper does not specify the exact testing configuration, such as image resolution, context length, cache size, etc. Here, we provide a direct comparison between BlueLM-V-3B and the statistics reported in the MiniCPM-V paper~\cite{yao2024minicpm}, as shown in Tab.~\ref{tab:deploy_minicpm}. We utilize an image with a resolution of 768$\times$1536, fix the whole input token length to 2048, and set the KV cache length to 2048. BlueLM-V-3B achieves shorter latency and faster token throughput.

\section{Conclusion and Future Discussions}\label{conclusion}

This paper introduces BlueLM-V-3B, an algorithm and system co-design approach tailored for MLLMs on mobile platforms. Our goal is to maximize usability and performance while ensuring a positive user experience. We emphasize algorithm-system co-design and hardware-aware optimization for training MLLMs and deploying them on mobile phones. Extensive experiments and statistical analyses demonstrate the strong capability and high efficiency of BlueLM-V-3B on mobile devices. Future work will focus on optimizing the scalability of BlueLM-V-3B for a broader range of mobile devices and exploring advanced algorithms to enhance performance and usability. We hope our work will contribute to the advancement of this field of study.

{
    \small
    \bibliographystyle{ieeenat_fullname}
    \bibliography{main}
}

\clearpage
\section{Relaxed Aspect Ratio Matching}

\begin{minipage}{\textwidth}
\begin{algorithm}[H]
\caption{Relaxed Aspect Ratio Matching}
\begin{algorithmic}[1]
\Function{Relaxed\_Aspect\_Ratio\_Matching}{original\_size: $(W_{\text{orig}},H_{\text{orig}})$, possible\_ratios: List of $(m,n)$}
    \State \textbf{Initialize:} best\_fit $\gets$ None, $R_{e, \text{max}} \gets 0$, $R_{w, \text{min}} \gets \infty$
    \State $(W_{\text{orig}}, H_{\text{orig}}) \gets \text{original\_size}$

    \For{each $(m,n)$ in possible\_ratios}
        \State $(W, H) \gets (384 \times m, 384 \times n)$
        \State $\text{scale} \gets \min \left(\frac{W}{W_{\text{orig}}}, \frac{H}{H_{\text{orig}}}\right)$
        \State $\delta W \gets \text{int}(W_{\text{orig}} \cdot \text{scale})$
        \State $\delta H \gets \text{int}(H_{\text{orig}} \cdot \text{scale})$
        \State $R_e \gets \min(\delta W \cdot \delta H, W_{\text{orig}} \cdot H_{\text{orig}})$
        \State $R_w \gets W \cdot H - R_e$

        \If{\textcolor{red}{$(R_e - R_{e, \text{max}}) > \alpha \cdot R_{e, \text{max}}$ \textbf{or} 
            $\left((R_{e, \text{max}} - R_e) < \alpha \cdot R_{e, \text{max}} \;\text{and}\; R_w < R_{w, \text{min}}\right)$}}
            \State $R_{e, \text{max}} \gets R_e$
            \State $R_{w, \text{min}} \gets R_w$
            \State best\_fit $\gets (m,n)$
        \EndIf
    \EndFor

    \State \Return best\_fit
\EndFunction
\end{algorithmic}
\label{alg:pseudocode}
\end{algorithm}
\vspace*{\fill}
\end{minipage}

\hspace{-1.2em}To further expand the content of the main text, here we provide more about the relaxed aspect ratio matching method in this section.\\

\hspace{-1.2em}\textbf{Pseudocode:} We present the pseudocode for our proposed relaxed aspect ratio matching method, as shown in Alg.~\ref{alg:pseudocode}. To be specific, we change the updating logic of LLaVA-NeXT by adding a parameter 
$\alpha$ such that when:
\begin{equation}
R_e - R_{e, \text{max}} > \alpha \cdot R_{e, \text{max}},
\end{equation}
or
\begin{equation}
(R_{e, \text{max}} - R_e) < \alpha \cdot R_{e, \text{max}}\; \text{and} \; R_w < R_{w, \text{min}},
\end{equation}
we then update
\begin{equation}
R_{e, \text{max}} \leftarrow R_e, \quad R_{w, \text{min}} \leftarrow R_w,
\end{equation}
and record the according aspect ratio. This increases the likelihood of selecting aspect ratios with smaller \( R_e \) but also smaller \( R_w \).\\
\begin{figure}
    \centering
    \vspace{20em}
    \includegraphics[width=\columnwidth]{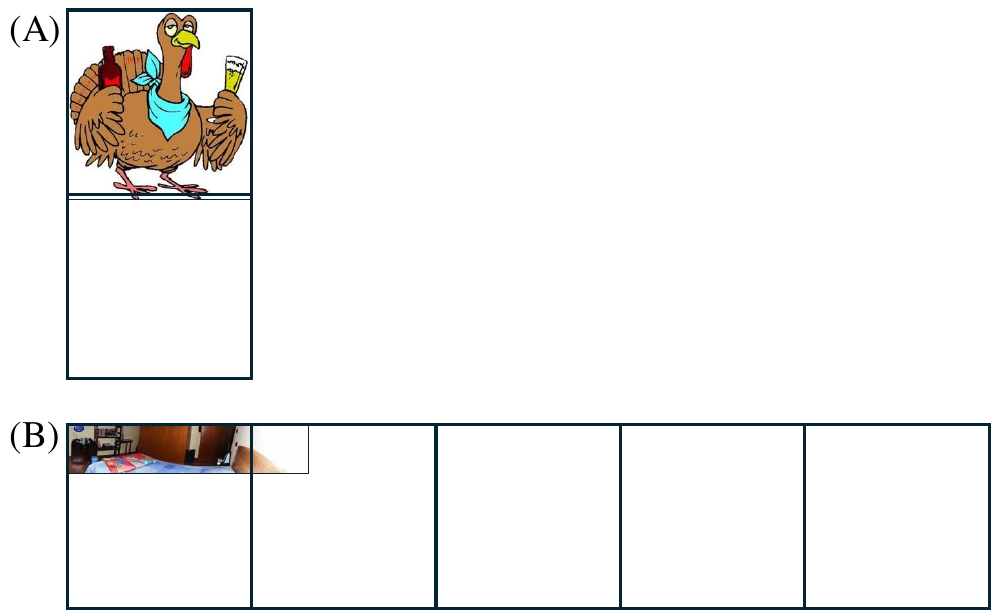}
    \caption{\textbf{Case study.} (A) LLaVA-NeXT chooses resolution 384$\times$768 for an image with the original size of 380$\times$393. (B) InternVL 1.5 chooses resolution 1920$\times$384 for an image with the original size of 500$\times$102.}
    \label{fig:big_case}
\end{figure}\\
\hspace{-1.2em}\textbf{Case Study:} Here we present real cases where LLaVA-NeXT~\cite{liu2024llavanext} and InternVL 1.5~\cite{chen2024far} result in significant image enlargement, as illustrated in Fig.~\ref{fig:big_case}. In Fig.~\ref{fig:big_case}A (from COYO-300M Dataset~\cite{kakaobrain2022coyo-700m}), LLaVA-NeXT selects a resolution of 384$\times$768 for an image originally sized 380$\times$393. Similarly, Fig.~\ref{fig:big_case}B (from CogVLM-SFT dataset~\cite{wang2023cogvlm}) shows InternVL 1.5 selecting a resolution of 1920$\times$384 for an image initially sized 500$\times$102. In contrast, our proposed relaxed aspect ratio matching selects 384$\times$384 for the 380$\times$393 image and 768$\times$384 for the 500$\times$102 image.
\clearpage
\section{Open-source Training Dataset}

Here we provide the open-source dataset used to train BlueLM-V-3B in the fine-tuning stage in Tab.~\ref{tab:data}.\\
\vspace{-1em}
\begin{table}[H]
\vspace{-1em}
\begin{tabular}{>{\raggedright\arraybackslash}p{0.9cm}>{\raggedright\arraybackslash}p{6.6cm}} %
\midrule
\multicolumn{1}{l}{\textbf{Task}} & \multicolumn{1}{c}{\textbf{Dataset}} \\ 
\midrule
{\textbf{Text-only}} & {ALLaVA~\cite{chen2024allava},   ScienceQA~\cite{lu2022learn}, Orca-Math~\cite{mitra2024orca}, OpenOrca~\cite{OpenOrca}, MetaMathQA~\cite{yu2023metamath}, WizardLM~\cite{xu2023wizardlm}, MathInstruct~\cite{toshniwal2024openmathinstruct}} \\ \midrule
{\textbf{Caption}} & {TextCaps~\cite{sidorov2020textcaps},   Screen2Words~\cite{wang2021screen2words}, VizWiz~\cite{gurari2020captioning}, Laion~\cite{schuhmann2022laion}, COCO~\cite{chen2015microsoft}, LLaVA~\cite{liu2023llava}, ALLaVA~\cite{chen2024allava}, SVIT~\cite{zhao2023svit}, SA1B~\cite{kirillov2023segment}, VSR~\cite{liu2023visual},   Chart2Text~\cite{kantharaj2022chart}, MultiMath~\cite{peng2024multimath}, ArXivCap~\cite{li2024multimodal}}, COYO~\cite{kakaobrain2022coyo-700m} \\ \midrule
{\textbf{OCR}} & {Wukong~\cite{gu2022wukong},   HierText~\cite{long2022towards}, TextOCR~\cite{singh2021textocr}, WildReceipt~\cite{sun2021spatial}, DocILE~\cite{vsimsa2023docile}, SVRD~\cite{yu2023icdar}, DocLayNet~\cite{pfitzmann2206doclaynet}, XFUND~\cite{xu2022xfund}, COCO-Text~\cite{veit2016coco}, SROIE~\cite{huang2019icdar2019},   FUNSD~\cite{jaume2019funsd}, CORD~\cite{park2019cord}, Paper2Fig100k~\cite{rodriguez2023ocr}, Docmatix~\cite{laurençon2024building}, LAION-2B-OCR~\cite{lin2025parrot}, SynthDoG~\cite{kim2022donut}, WebSight~\cite{laurençon2024unlocking}, DeepForm~\cite{svetlichnaya2020deepform}, Kleister~\cite{stanislawek2021kleister}, TabFact~\cite{2019TabFactA}} \\ \midrule
{\textbf{VQA}} & {LVIS-Instruct4V~\cite{wang2023instruct4v},  CLEVR~\cite{johnson2017clevr}, TallyQA~\cite{acharya2019tallyqa}, LNQA~\cite{pont2020connecting}, Geo170K~\cite{sharma2024geocoder}, ALLaVA~\cite{chen2024allava},  DocVQA~\cite{mathew2021docvqa}, ChartQA~\cite{masry2022chartqa}, ArxivQA~\cite{li2024multimodal}, GEOS~\cite{seo2015solving}, PMC-VQA~\cite{zhang2023pmc}, KVQA~\cite{shah2019kvqa}, Geometry3K~\cite{lu2021inter}, MapQA~\cite{chang2022mapqa}, PlotQA~\cite{methani2020plotqa}, ViQuAE~\cite{lerner2022viquae}, VQA-RAD~\cite{lau2018dataset}, ST-VQA~\cite{biten2019scene}, TextVQA~\cite{singh2019textvqa}, LLaVAR~\cite{zhang2023llavar}, SIBR~\cite{yang2023modeling}, MMC-Inst~\cite{liu2023mmc}, IconQA~\cite{lu2021iconqa}, GQA~\cite{hudson2019gqa}, SciGraphQA~\cite{li2023scigraphqa},  LRV-Instruction~\cite{liu2023aligning}, DVQA~\cite{kafle2018dvqa}, InfographicVQA~\cite{mathew2022infographicvqa}, FigureQA~\cite{kahou2017figureqa}, WikiTableQuestions~\cite{pasupat2015compositional}, TAT-DQA~\cite{zhu2022towards},  VisualMRC~\cite{tanaka2021visualmrc}, ScienceQA~\cite{lu2022learn}, OCR-VQA~\cite{mishra2019ocrvqa}, WebSRC~\cite{chen2021websrc}, PathVQA~\cite{he2020pathvqa}, UniGeo~\cite{chen2022unigeo}, ScreenQA~\cite{hsiao2022screenqa}, VizWiz~\cite{gurari2018vizwiz}, SVIT~\cite{zhao2023svit}, CogVLM~\cite{wang2023cogvlm}, FM-IQA~\cite{gao2015you}, VQAv2~\cite{goyal2017making}, OK-VQA~\cite{marino2019ok}, EST-VQA~\cite{wang2020general}, VisDial~\cite{das2017visual}, Shikra~\cite{chen2023shikra}, Super-CLEVR~\cite{li2023super}, LLaVA~\cite{liu2023improvedllava}, IDK~\cite{cha2024visually}, AlfWorld~\cite{ALFWorld20}, M-HalDetect~\cite{gunjal2024detecting}, Cambrian7M~\cite{tong2024cambrian}, LLaVA-OneVision~\cite{li2024llava}, mPLUG-DocOwl~\cite{ye2023mplug}, UReader~\cite{ye2023ureader}} \\ \midrule
\end{tabular}%
\caption{\textbf{Training data.} This table presents the open-source datasets used in the fine-tuning stage, corresponding with the categories and data volume in Tab.~\ref{tab:data_ratio} of the main text.}
\label{tab:data}
\end{table}

\vspace{-1em}
\hspace{-1.2em}Please note that some datasets may belong to more than one category, and there may be overlapping data among these datasets.

\section{Hyper-parameters for Training}

We list the hyper-parameters for the pre-training stage (stage 1) and fine-tuning stage (stage 2) in Tab.~\ref{tab:stage1_param} and Tab.~\ref{tab:stage2_param} respectively.

\begin{table}[h]
\resizebox{\columnwidth}{!}{%
\begin{tabular}{lc}
\midrule
\textbf{Configuration} & \textbf{Stage   1} \\ \midrule
\textbf{LLM Sequence Length} & 4096 \\ \midrule
\textbf{Dynamic Resolution} & None (384$\times$384) \\ \midrule
\textbf{Optimizer} & AdamW \\ \midrule
\textbf{Optimizer Hyperparams} & $\beta_1=0.9$, $\beta_2=0.98$,   $\epsilon=10^{-6}$ \\ \midrule
\textbf{Peak LR} & $10^{-3}$ \\ \midrule
\textbf{LR Schedule} & Cosine Decay \\ \midrule
\textbf{Weight Decay} & 0.05 \\ \midrule
\textbf{Training Steps} & 3.434k \\ \midrule
\textbf{Warm-up Steps} & 34 \\ \midrule
\textbf{Global Batch Size} & 720 \\ \midrule
\textbf{Gradient Accumulation} & 1 \\ \midrule
\textbf{Numerical Precision} & \texttt{bfloat16} \\ \midrule
\end{tabular}%
}
\caption{\textbf{Hyper-parameters.} Hyper-parameters for the pre-training stage (stage 1).}
\label{tab:stage1_param}
\end{table}

\begin{table}[h]
\resizebox{\columnwidth}{!}{%
\begin{tabular}{lc}
\midrule
\textbf{Configuration} & \textbf{Stage   2} \\ \midrule
\textbf{LLM Sequence Length} & 4096 \\ \midrule
\textbf{Dynamic Resolution} & Up to 16 patches (1536$\times$1536) \\ \midrule
\textbf{Optimizer} & AdamW \\ \midrule
\textbf{Optimizer Hyperparams} & $\beta_1=0.9$, $\beta_2=0.98$,   $\epsilon=10^{-6}$ \\ \midrule
\textbf{Peak LR} & $10^{-4}$ \\ \midrule
\textbf{LR Schedule} & Cosine Decay \\ \midrule
\textbf{Weight Decay} & 0.05 \\ \midrule
\textbf{ViT Layer-wise LR Decay} & 0.9 \\ \midrule
\textbf{Training Steps} & 131k \\ \midrule
\textbf{Warm-up Steps} & 1310 \\ \midrule
\textbf{Global Batch Size} & 5760 \\ \midrule
\textbf{Gradient Accumulation} & 8 \\ \midrule
\textbf{Numerical Precision} & \texttt{bfloat16} \\ \midrule
\end{tabular}%
}
\caption{\textbf{Hyper-parameters.} Hyper-parameters for the fine-tuning stage (stage 2).}
\label{tab:stage2_param}
\end{table}

\hspace{-1.2em}Please note that due to the upsampling of certain datasets with smaller data volumes, the product of Training Steps and Global Batch Size may exceed the total data volume.

\clearpage
\section{Visualization on Practical Cases}

\begin{minipage}{\textwidth}
We present some demos of BlueLM-V-3B on mobile devices.\\
\vspace{1em}
\end{minipage}

\begin{minipage}{\textwidth}
\begin{figure}[H]
    \centering
    \vspace{-2em}
    \includegraphics[width=0.7\linewidth]{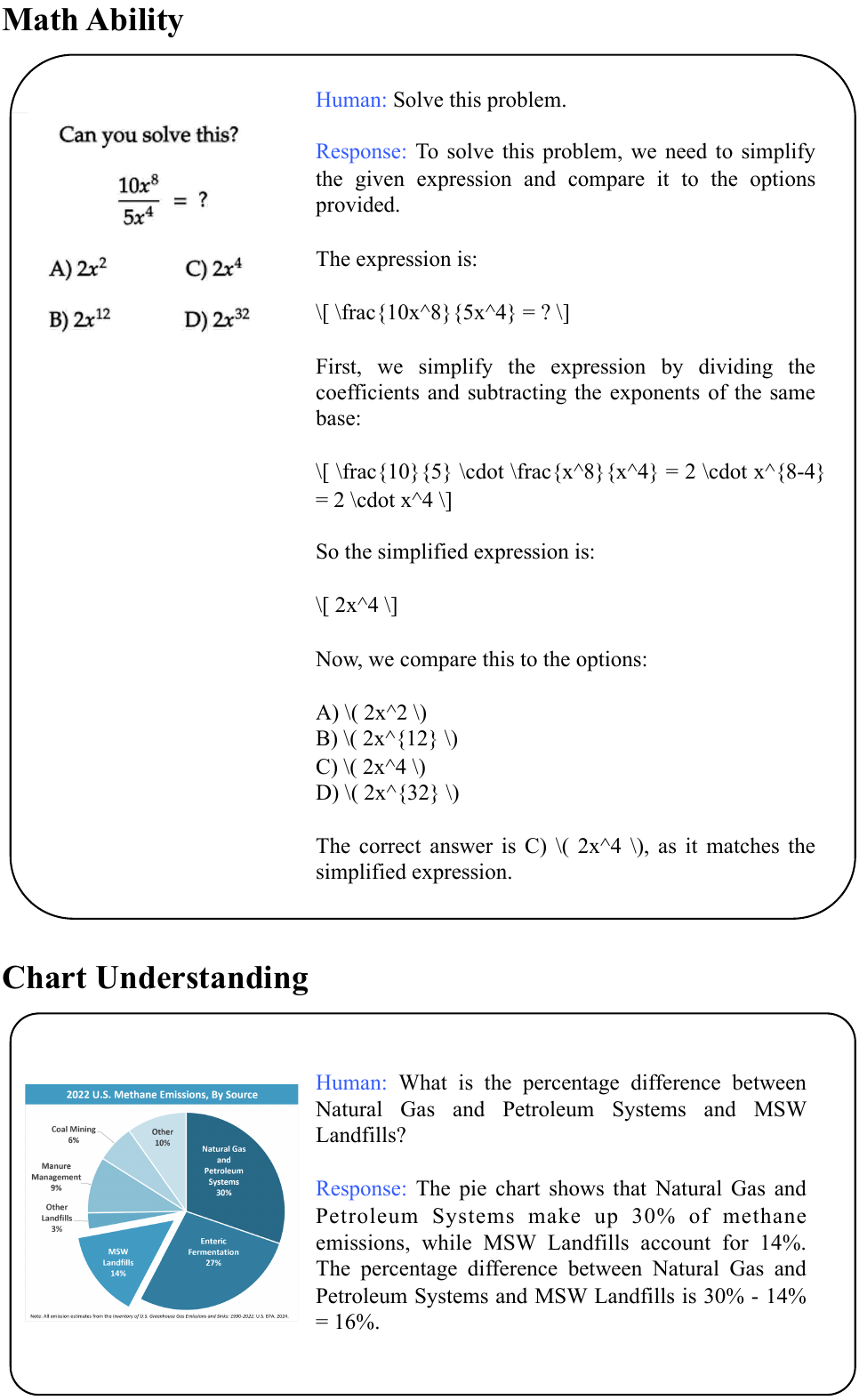}
    \label{fig:demo1}
\end{figure}
\end{minipage}

\clearpage
\begin{minipage}{\textwidth}
\begin{figure}[H]
    \centering
    \vspace{-2em}
    \includegraphics[width=0.7\linewidth]{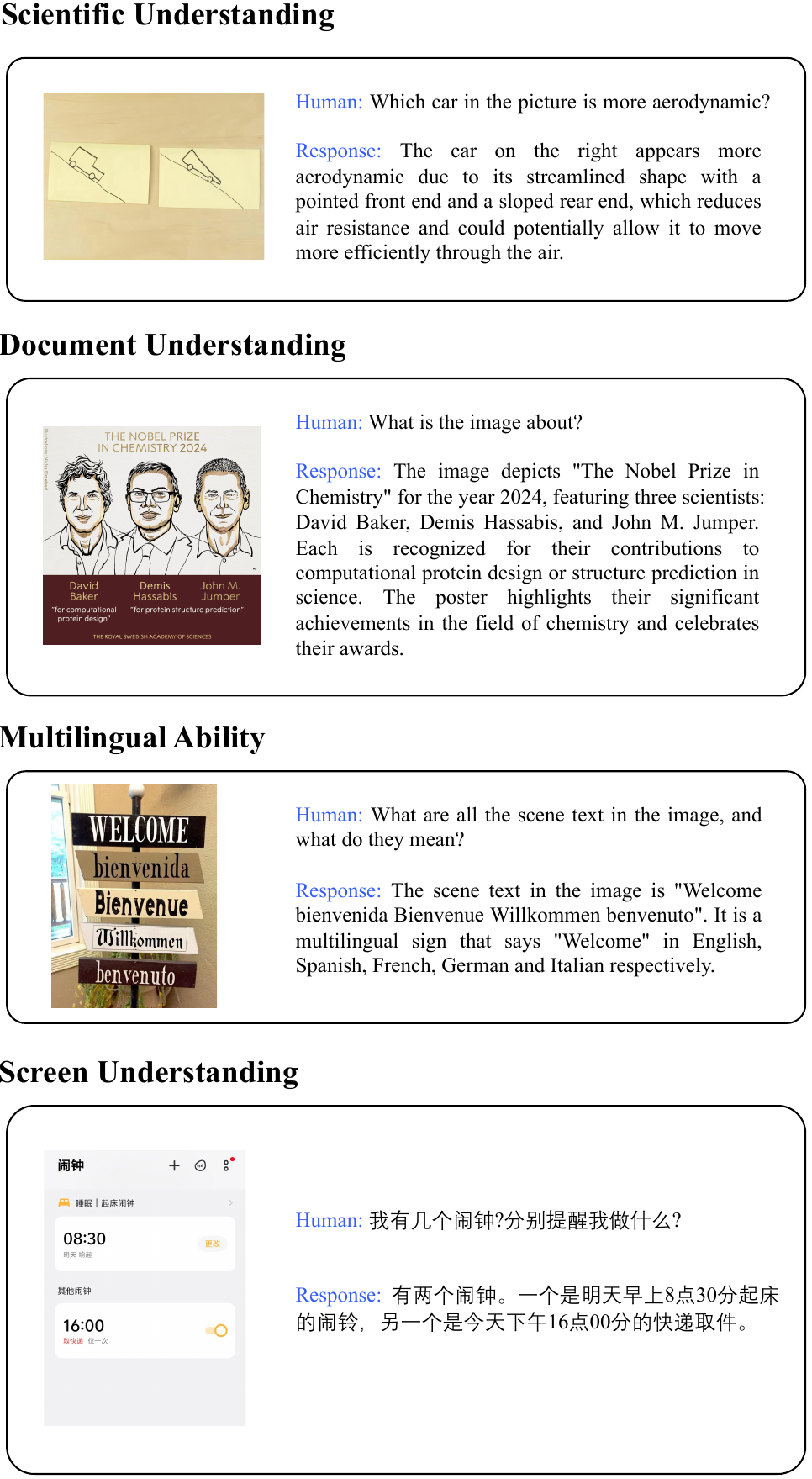}
    \label{fig:demo2}
\end{figure}
\end{minipage}

\clearpage
\begin{minipage}{\textwidth}
\begin{figure}[H]
    \centering
    \vspace{-2em}
    \includegraphics[width=0.7\linewidth]{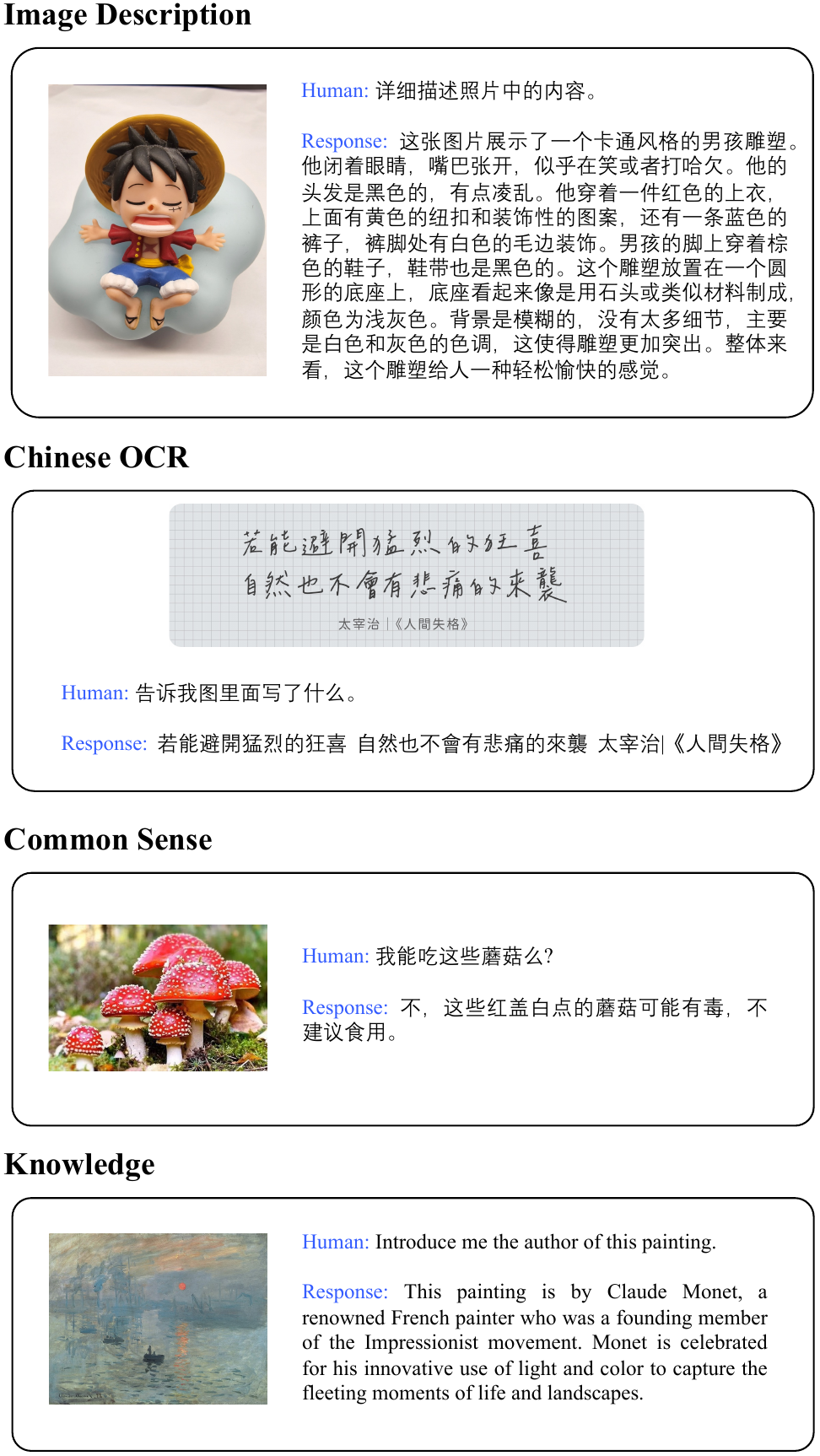}
    \label{fig:demo3}
\end{figure}
\end{minipage}

\clearpage
\begin{minipage}{\textwidth}
\begin{figure}[H]
    \centering
    \vspace{-2em}
    \includegraphics[width=0.7\linewidth]{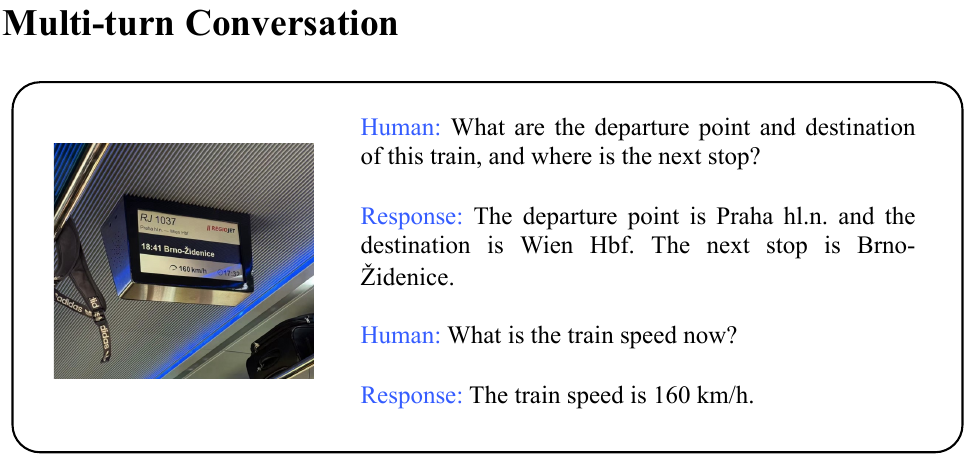}
    \label{fig:demo4}
\end{figure}
\end{minipage}

\end{document}